%% file: pplr.tex
\documentclass{sig-alternate}

\usepackage{cite}
\usepackage{graphicx}
\usepackage{amsmath,amssymb}
\usepackage{url}

\begin{document}

\title{Privacy-Preserving Spam Filtering}

\numberofauthors{3}

\author{
  \alignauthor Manas A. Pathak \\
  \affaddr{Carnegie Mellon University} \\
  \email{manasp@cs.cmu.edu}
  \alignauthor Mehrbod Sharifi \\
  \affaddr{Carnegie Mellon University} \\
  \email{mehrbod@cs.cmu.edu}
  \alignauthor Bhiksha Raj \\
  \affaddr{Carnegie Mellon University} \\
  \email{bhiksha@cs.cmu.edu}
}

\maketitle

\begin{abstract}
  \input{abstract}
\end{abstract}

\terms{Privacy Preserving Machine Learning, Spam Filtering}

\section{Introduction}
\input{introduction}

\input{related-work}

\section{Preliminaries}
\input{lr}
\input{homomorphic-encryption}

\input{protocol}

\input{analysis}

\input{experiments}

\input{conclusion}

\bibliographystyle{abbrv}
\bibliography{pplr}

\end{document}

%% file: abstract.tex
Email is a private medium of communication, and the inherent privacy
constraints form a major obstacle in developing effective
spam filtering methods which require access to a large
amount of email data belonging to multiple users. To mitigate this
problem, we envision a privacy preserving spam filtering system, where
the server is able to train and evaluate a logistic regression based
spam classifier on the combined email data of all users without being
able to observe any emails using primitives such as homomorphic
encryption and randomization. We analyze the protocols for correctness
and security, and perform experiments of a prototype system on a large
scale spam filtering task. 

State of the art spam filters often use character n-grams as features
which result in large sparse data representation, which is not
feasible to be used directly with our training and evaluation protocols. We
explore various data independent dimensionality reduction which
decrease the running time of the protocol making it feasible to use in
practice while achieving high accuracy.

%% file: introduction.tex
Email is a private medium of communication with the message
intended to be read only by the recipients. Due to the sensitive nature
of the information content, there might be personal, strategic,
and legal constraints against sharing and releasing email data. These
constraints form formidable obstacles in many email processing applications
such as spam filtering which are usually supplied by a separate
service provider.

Over the years, spam has become a major problem: 75.9\% all
  emails sent in August 2011 were spam~\cite{spam-report}. Email users
can benefit from using accurate spam filters, which could greatly
reduce the loss of time and productivity due to spam email.  A
proficient user can directly learn a spam filtering classifier on her
own private data and send it to the spam filtering provider or apply
it herself, diminishing the need for a privacy preserving spam
filtering system. It is, however, seen that the accuracy of spam filters
based on classification models can be vastly improved by training on
aggregates of data obtained from a large number of email users. This
training and application of spam filters should, however, not be at
the expense of user privacy, with users being required to make their
emails available to the spam filtering service provider.

In this paper we propose a solution that enables users to share their
private email data to train and apply spam filters while satisfying
privacy constraints. We choose \emph{logistic regression} as our
classification model as it is widely used in spam filtering and text
classification applications and is observed to achieve very high accuracy
in these tasks. The training algorithm for logistic regression based
on gradient ascent is more amenable to be modified to satisfy privacy
constraints. The update step in the training algorithm is also
particularly convenient when the training data is split among \emph{multiple
parties}, who can simply compute their gradient on their private email
data and the server can privately aggregate these to update the model
parameters. Furthermore, logistic regression is can also be easily
modified to the \emph{online learning} setting. In a practical spam filtering
system, as the users are unlikely to relabel previously read emails as spam,
the classifier needs to be learned on a continuously arriving stream
of email data.

Although primarily directed at spam filtering, our solution can also
be applied to any form of private text classification and in general
to any binary classification setting where privacy is important,
\emph{e.g.}, predicting the likelihood of disease based on an individual's
private medical records. Our methods also extend to batch processing
scenarios.

Formally, we consider two kinds of parties: a set of users who have
access to their private emails and server who is interested in
training a spam classification model over the complete email data.
The users can communicate with the server but not with each other as
this is typically the case in an email service.  The primary privacy
constraint is that the server should not be able to observe the emails
belonging to any of the users and similarly, any user should not be
able to observe emails belonging to any other user. The secondary
privacy constraint is that the users should not be able to observe the
parameters of the classification model learned by the server. While
the motivation behind the former privacy constraint is more obvious,
the server might want to keep the classification model private if it
was privately trained over large quantities of training data pooled
from a large number of users and if the server is interested in
offering a restricted pay per use spam filtering service. We present
protocols to train and evaluate logistic regression models while
maintaining these privacy constraints.

Our privacy preserving protocol falls into the broad class of
\emph{secure multiparty computation} (SMC) algorithms~\cite{Yao82}. In
the SMC framework, multiple parties desire to compute a function that
combines their individual inputs. The privacy constraint is that no
party should learn anything about inputs belonging to any other party
besides what can be inferred from the value of the function. We
construct our protocol using a cryptosystem satisfying
\emph{homomorphic encryption}~\cite{Paillier99}, in which operations
on encrypted data correspond to operations on the original unencrypted
data (Section~\ref{sec:homomorphic-encryption}). We further augment
our protocol with additive and multiplicative randomization, and
present an information theoretic analysis of the security of the
protocol.

The benefit of training and evaluating a spam filtering classifier
privately comes with a substantial overhead of computation and data
transmission costs. We find that these costs are linear in the number
of training data instances and the data dimensionality. As the size of
our character four-gram feature representation of the text data is
extremely large (\emph{e.g.}, one million features), application of
our protocol on a typical email dataset is prohibitively
expensive. Towards this, we apply suitable data dimensionality
reduction techniques to make the training protocol computationally
usable in practical settings. As the same dimensionality reduction is
has to be applied by all the parties to their private data, we require
that the techniques used are data independent and do not require to be
computed separately. We present extensive evaluation of our protocol
on a large scale email dataset from the CEAS 2008 spam filtering
challenge. With data independent dimensionality reduction such as
locality sensitive hashing, multinomial sampling, and hash space
reduction, we demonstrate that our protocol is able to achieve state
of the art performance in a feasible amount of running time.

To summarize, our main contributions are:
\begin{itemize}
\item Protocols for training and evaluating the logistic regression
  based spam filtering classifier with online updates from while
  preserving the private email data belonging to multiple parties.
\item Analysis of the protocols for security and efficiency.
\item Dimensionality reduction for making the protocol feasible to be
  used in a practical spam filtering task.
\item Experiments with the privacy preserving training and evaluation
  protocols over a large scale spam dataset: trade off between
  running time and accuracy.
\end{itemize}

%% file: related-work.tex
\section{Related Work}

Email spam filtering is a well established area of research. The
accuracy of the best systems in the 2007 CEAS spam filtering
competition was better than 0.9999~\cite{Cormack07}. Our
implementation is an online logistic regression classifier
implementation inspired by \cite{Goodman06} which on application to
binary character four-gram features was shown to have near state of
the art accuracy~\cite{Cormack07}. 

The application of privacy preserving techniques to large scale real
world problems of practical importance, such as spam filtering, is an
emerging area of research.  Li, et al.~\cite{Li2009} present a
distributed framework for privacy aware spam filtering. Their method
is based on applying a one-way fingerprinting
transformation~\cite{shingles} to the message text and comparing two
emails using a Hamming distance metric and does not involve
statistical learning. Additionally, this method also requires that the
spam emails belonging to all users should be revealed which does not
match our privacy criteria. We consider all emails to be
private as the nature of the spam emails a user receives
might be correlated to the user's online and offline activities.

There has also been recent work on constructing privacy preserving
protocols for general data mining tasks including decision
trees~\cite{VaidyaCKP08}, 
clustering~\cite{LinCZ05}, naive Bayes~\cite{VaidyaKC08}, and support
vector machines~\cite{jvppsvm}. To the best of our knowledge, this
paper is the first to describe a practical privacy-preserving
framework using a logistic regression classifier applied to a real
world spam filtering task.

%% file: lr.tex
\subsection{Classification Model: Logistic Regression in the Batch and
  Online Settings}\label{sec:lr} 

The training dataset consisting of $n$ documents classified by the user
as spam or ham (\emph{i.e.}, not spam) are represented as the labeled
data instances $(x,y)=\{(x_1,y_1),\ldots,(x_n,y_n)\}$ where $x_i \in
\mathbb{R}^d$ and $y_i \in \{-1,1\}$. In the batch learning setting,
we assume that the complete dataset is available at a given time.
In the logistic regression classification algorithm, we model the
class probabilities by a sigmoid function
\[ P(y_i = 1 | x_i,w) = \frac{1}{1+e^{-y_iw^Tx_i}}. \]
We denote the log-likelihood for the weight vector $w$ computed
over the data instances $(x,y)$ by $L(w,x,y)$.
Assuming the data instances to be i.i.d., the data log-likelihood
$L(w,x,y)$ is equal to 
\[  L(w,x,y) = \log \prod_i \frac{1}{1+e^{-y_iw^Tx_i}} = -\sum_i\log[1+e^{-y_iw^Tx_i}]. \]
We maximize the data log-likelihood $L(w,x,y)$ using gradient ascent
to obtain the classifier with the optimal weight vector
$w^*$. Starting with a uniformly initialized vector $w_{(0)}$, in
the $t^{\rm th}$ iteration, we update $w_{(t)}$ as
\begin{align}
  w_{(t+1)} = w_{(t)} + \eta \nabla L(w_{(t)},x,y)
  = w_{(t)} + \eta \sum_i \frac{y_ix_i^T}{1+e^{y_iw_{(t)}^Tx_i}},
\end{align}
where $\eta$ is the pre-defined step size. We terminate the procedure
on convergence between consecutive values of $w_{(t)}$.

In the online learning setting, the data instances are obtained
incrementally rather than being completely available at a given
instance of time. In this case, we start with a model with the uniformly random
weight vector $w_{(0)}$. A model $w_{(t)}$ learned using the first $t$
instances, is updated after observing a small block of $k$ instances.
with the gradient of the log-likelihood computed over that block.

%% file: homomorphic-encryption.tex
\subsection{Homomorphic Encryption} \label{sec:homomorphic-encryption}

In a homomorphic cryptosystem, operations performed on encrypted data
(\emph{ciphertext}) map to the corresponding operations performed on the
original unencrypted data (\emph{plaintext}). If $+$ and $\cdot$ are two
operators and $x$ and $y$ are two plaintexts, a homomorphic encryption
function $E$ satisfies
\[ E[x] \cdot E[y] = E[x + y]. \]
This allows one party to encrypt the data using a homomorphic
encryption scheme and another party to perform operations without
being able to observe the plaintext data. This property forms the
fundamental building block of our privacy preserving protocol.

In this work we use the additively homomorphic Paillier
cryptosystem~\cite{Paillier99} which also satisfies semantic security. The
Paillier key generation algorithm produces a pair of $b$-bit numbers
$(N,g)$ constituting the public key corresponding to the encryption
function $E : \mathbb{Z}_N \mapsto \mathbb{Z}_{N^2}$ and another pair
of $b$-bit numbers $(\lambda,\mu)$ constituting the private key
corresponding to the decryption function $D : \mathbb{Z}_{N^2} \mapsto
\mathbb{Z}_N$.

Given a plaintext $x \in \mathbb{Z}_N$, the encrypted text is given by:
\[ E[x] = g^x r^n \mod N^2, \]
where $r$ is a random number sampled uniformly from
$\mathbb{Z}_N$. Using a different value of the random number $r$
provides semantic security, \emph{i.e.}, two different encryptions of a number
$x$ say, $E[x;r_1]$ and $E[x;r_2]$ will have different values but
decrypting each of them will result in the same number $x$. It can be
easily verified that the above encryption function satisfies the
following properties:
\begin{enumerate}
\item For any two ciphertexts $E[x]$ and $E[y]$, 
  \[ E[x] ~ E[y] = E[x + y\mod N^2]. \]
\item And as a corollary, for any ciphertext $E[x]$ and plaintext $y$, 
  \[ E[x]^y = E[x ~ y\mod N^2]. \]
\end{enumerate}

\subsubsection*{Extending the Encryption Function to Real Numbers}

Paillier encryption as most other cryptosystems is defined over the
finite field $\mathbb{Z}_N=\{0,\ldots,N-1\}$. However, in our protocol
we need to encrypt real numbers, such as the training data and model
parameters. We make the following modifications to the encryption
function to support this.
\begin{enumerate}
  \item Real numbers are converted to a fixed precision floating point
    representation. For a large constant $C$, a real number $x$ is
    represented as $\lfloor Cx \rfloor=\bar{x}$.
    \[ E[\bar{x}] = E[\lfloor Cx \rfloor], \quad
       D[E[\bar{x}]] = E[\lfloor Cx \rfloor]/C = x. \] 

  \item The encryption of a negative integer is represented by the
    encryption of its modular additive inverse. If $-x$ is a negative
    integer,
    \[ E[-x] = E[N-x]. \]

  \item Exponentiation of an encrypted number by a negative integer is
    represented as the exponentiation of the multiplicative inverse of
    the encryption in the $\mathbb{Z}_{N^2}$ field, by the
    corresponding positive integer. We represent the
    exponentiation\footnote{We slightly abuse the notation to
      represent the non-modular exponentiation of the ciphertext by
      $E[x]^{a}$ to refer to $E[x] \cdot E[x] \cdots $ ($a$ times).}
    the ciphertext $E[x]$ by a negative integer $-y$ as 
    \[ E[x]^{-y} = E[x^{-1} \mod N^2]^y. \]    
\end{enumerate}

Representing real numbers by a fixed precision number 
introduces a small error due to the truncation which is
directly proportional to the value of $C$. This representation also 
reduces the domain of the encryption function from
$\{0,\ldots,N-1\}$ to $\{0,\ldots,\lfloor\frac{N-1}{C}\rfloor\}$. 
We need to ensure that the result of homomorphic operations on encrypted
functions do not overflow the range, so we need to increase the
bit-size $b$ of the encryption keys proportionally with $C$. As the
computational cost of the encryption operations is also proportional
to $b$, this creates a trade-off between accuracy and computation
cost. 

The representation of negative integers on the other hand does not
introduce any error but further halves the domain of the encryption
function from $\{0,\ldots,\lfloor\frac{N-1}{C}\rfloor\}$ to
$\{0,\ldots,\lfloor\frac{N-1}{2C}\rfloor\}$ which we denote by $\mathcal{D}$.

%% file: protocol.tex
\section{Privacy Preserving Classifier \\ Training and Evaluation}\label{sec:protocol}

\subsection{Data Setup and Privacy Conditions}

We define the party ``Bob'' who is interested in training a logistic
regression classifier with weight vector $w\in\mathbb{R}^d$. In the
online learning setting, multiple users interact with Bob at one time
using their private training data as input. As all these
parties play the same role in their interactions with Bob in one
update step, we represent them by a generic user ``Alice''.  
Later on we see how Bob privately aggregates the encrypted gradients
provided by individual parties. 

Alice has a sequence of labeled training data instances
$(x,y)=\{(x_1,y_1),\ldots,(x_n,y_n)\}$. Bob is interested in training
a logistic regression classifier with weight vector $w\in\mathbb{R}^d$
over $(x,y)$ as discussed in Section~\ref{sec:lr}.  The privacy
constraint implies that Alice should not be able to observe $w$ and
Bob should not be able to observe $(x_i,y_i)$.  The parties are
assumed to be \emph{semi-malicious}, \emph{i.e.}, they correctly
execute the steps of the protocol and do not attempt to cheat by using
fraudulent data as input in order to extract additional information
about the other parties. The parties are assumed to be curious,
\emph{i.e.}, they keep a transcript of all intermediate results and
can use that to gain as much information as possible.

\subsection{Private Training Protocol}\label{sec:training-protocol}

Bob generates a public and private key pair for a $b$-bit Paillier
cryptosystem and provides the public key to Alice. In this
cryptosystem, Bob is able to perform both encryption and decryption
operations while Alice can perform only encryption.

As mentioned before, we use the homomorphic properties of Paillier
encryption to allow the parties to perform computations using private
data. The update rule requires Bob to compute the gradient of
the data log-likelihood function $\nabla L(w_{(t)},x,y)$ which
involves exponentiation and division and cannot be done using only
homomorphic additions and multiplications. We supplement the
homomorphic operations with Bob performing those operations on
multiplicative shares to maintain the privacy constraints. 
As mentioned in Section~\ref{sec:homomorphic-encryption}, the domain
of the encryption function is
$\mathcal{D}=\{0,\ldots,\lfloor\frac{N-1}{2C}\rfloor\}$.
We sample the randomizations uniformly from this set.

Bob initiates the protocol with a uniform $w_{(0)}$ and the gradient update
step $\eta$ is publicly known. We describe the $t^{\rm th}$ iteration
of the protocol below.~\\

\noindent \textbf{Input:} Alice has $(x,y)$ and the encryption key, \\
Bob has $w_{(t)}$ and both encryption and decryption keys. \\
\textbf{Output:} Bob has $w_{(t+1)}$.
\begin{enumerate}
\item\label{step:init} Bob encrypts $w_{(t)}$ and transfers $E[w_{(t)}]$ to Alice.
\item For each training instance $x_i$, $i=1,\ldots,n$, Alice computes
  \[ \prod_{j=1}^d E[w_{(t)j}]^{y_ix_{ij}} 
  = E\left[\sum_{j=1}^d y_iw_{(t)j}x_{ij}\right] = E\left[y_i w_{(t)}^Tx_{i}\right]. \]
\item Alice samples $n$ numbers $r_1,\ldots,r_n$ uniformly from
  $\mathbb{Z}_N=\{1,\ldots,N-1\}$ and computes
  \[ E\left[y_i w_{(t)}^Tx_{i}\right] \cdot E[-r_i] = E\left[y_i
    w_{(t)}^Tx_{i} -r_i\right]. \]
  Alice transfers $E\left[y_i w_{(t)}^Tx_{i} -r_i\right]$ to Bob.
\item Bob decrypts this to obtain $y_i w_{(t)}^Tx_{i} -r_i$. In this
  way, Alice and Bob have additive shares of the inner products $y_i
  w_{(t)}^Tx_{i}$.

\item Bob exponentiates and encrypts his shares of the inner products.
  He transfers $E\left[e^{y_i w_{(t)}^Tx_{i}-r_i}\right]$ to
  Alice.

\item Alice homomorphically multiplies the quantities she obtained from Bob by the
  exponentiations of her corresponding random shares to obtain the
  encryption of the exponentiations of the inner products.\footnote{In
    some cases, the exponentiation might cause the 
    plaintext to overflow the domain of encryption function. This can
    be handled by computing the sigmoid function homomorphically using
    a piecewise linear sum of components.}
  \[ E\left[e^{y_i w_{(t)}^Tx_{i}-r_i}\right]^{e^{r_i}} 
     = E\left[e^{y_i w_{(t)}^Tx_{i}}\right]. \]
 Alice homomorphically adds $E[1]$ to these quantities to obtain
 $E\left[1+e^{y_i w_{(t)}^Tx_{i}}\right]$.

\item Alice samples $n$ numbers $q_1,\ldots,q_n$ from $\mathcal{D}$ using a
  bounded Power law distribution\footnote{We require that $q$ has the pdf 
    $P(q) \propto 1/q$ for $1 \le q \le |\mathcal{D}|$. $q$ can be
    generated using inverse transform sampling. We discuss the reasons
    for this in Section~\ref{sec:security}.}.
  She then homomorphically computes
  \[ E\left[1+e^{y_iw_{(t)}^Tx_{i}}\right]^{q_i} 
  = E\left[q_i\left(1+e^{y_iw_{(t)}^Tx_{i}}\right)\right]. \]
  She transfers these quantities to Bob.

\item Bob decrypts these quantities and computes the reciprocal
  $\frac{1}{q_i\left(1+e^{y_iw_{(t)}^Tx_{i}}\right)}$. He then
  encrypts the reciprocals and sends them to Alice.

\item Alice homomorphically multiplies $q_i$ with the encrypted
  reciprocals to cancel out her multiplicative share.
  \[ E\left[\frac{1}{q_i\left(1+e^{y_iw_{(t)}^Tx_{i}}\right)}\right]^{q_i}
     = E\left[\frac{1}{1+e^{y_iw_{(t)}^Tx_{i}}}\right]. \]

\item Alice then homomorphically multiplies the encrypted reciprocal
  by each component of $y_ix_i^T$ to obtain the encrypted
  $d$-dimensional vector
  \[ E\left[\frac{1}{1+e^{y_iw_{(t)}^Tx_{i}}}\right]^{y_ix_i^T} 
     = E\left[\frac{y_ix_i^T}{1+e^{y_iw_{(t)}^Tx_i}}\right]. \]
  She homomorphically adds each encrypted component to obtain
  \[ \prod_i E\left[\frac{y_ix_i^T}{1+e^{y_iw_{(t)}^Tx_{i}}}\right]
     = E\left[\sum_i
       \frac{y_ix_i^T}{1+e^{y_iw_{(t)}^Tx_{i}}}\right]. \]
  This is the encrypted gradient vector $E\left[\nabla L(w_{(t)},x,y)\right]$.

\item\label{step:update} Alice homomorphically updates the encrypted
  weight vector she obtained in Step \ref{step:init} with the
  gradient.
  \begin{align*}
    E[w_{(t+1)}] &= E[w_{(t)}] ~ E\left[\nabla L(w_{(t)},x,y)\right]^\eta \\
    &= E\left[w_{(t)} + \eta \nabla L(w_{(t)},x,y)\right].
  \end{align*}

\item Alice then sends the updated weight vector $E[w_{(t+1)}]$ to Bob who
  then decrypts it to obtain his output.
\end{enumerate}

In this way, Bob is able to update his weight vector using Alice's
data while maintaining the privacy constraints. In the batch setting, 
Alice and Bob repeat Steps 2 to 11 to perform the iterative gradient
descent. Bob can check for convergence in the value of $w$ between
iterations by performing Step 12.
In the online setting, Alice and Bob execute the protocol only once
with using Alice using a typically small block of $k$ data instances as
input.

\subsubsection*{Extensions to the Training Protocol}

\begin{enumerate}
\item \emph{Training on private data horizontally split across multiple parties.}

In the online setting we do not make any assumption about which data
holding party is participating in the protocol. Just as Alice uses her
data to update $w$ privately, other parties can then use their data to
perform the online update using the same protocol.

In the batch setting, multiple parties can execute one iteration of
the protocol individually with Bob to compute the encrypted gradient
on their own data. Finally, Bob can receive the encrypted gradients
from all the parties and update the weight vector as follows.
\[ w_{(t+1)} = w_{(t)} + \eta \sum_k \nabla L(w_{(t)},x^k,y^k), \]
where $(x^1,y^1),\ldots,(x^K,y^K)$ are the individual datasets
belonging to $K$ parties.

\item \emph{Training a regularized classifier.}

The protocol can easily be extended to introduce $\ell_2$
regularization, which is a commonly used method to prevent
over-fitting. In this case the update rule becomes
\[ w_{(t+1)} = w_{(t)} + \eta \nabla L(w_{(t)},x,y) + 2 \lambda w_{(t)}, \]
where $\lambda$ is the regularization parameter.

This can be accommodated by Alice homomorphically adding the term $2 \lambda
w_{(t)}$ to the gradient in Step 11.
\begin{align*}
  E[w_{(t+1)}] &= E[w_{(t)}]^{1+2\lambda} 
  ~ E\left[\nabla L(w_{(t)},x,y)\right]^\eta \\
  &= E\left[(1+2\lambda)w_{(t)} + \eta \nabla L(w_{(t)},x,y)\right].
\end{align*}

In order to identify the appropriate value of $\lambda$ to use, Alice
and Bob can perform $m$-fold cross-validation by repeatedly executing
the private training and evaluation protocols over different subsets
of data belonging to Alice.

\end{enumerate}

\subsection{Private Evaluation Protocol}\label{sec:evaluation-protocol}

Another party ``Carol'' having one test data instance $x' \in
\mathbb{R}^d$ is interested in applying the classification model with
weight vector $w$ belonging to Bob. Here, the privacy constraint require
that Bob should not be able to observe $x'$ and Carol should not be
able to observe $w$.
Similar to the training protocol, Bob generates a public and private
key pair for a $b$-bit Paillier cryptosystem and provides the public
key to Carol.

In order to label the data instance as $y'=1$, Carol
needs to check if $P(y'=1|x',w) = \frac{1}{1+e^{-w^Tx'}} >
\frac{1}{2}$ and vice-versa for $y'=-1$. This is equivalent to
checking if $w^Tx' > 0$. We develop the following protocol towards
this purpose.~\\

\noindent \textbf{Input:} Bob has $w$ and generates a public-private
key pair. \\
Carol has $x'$ and Bob's public key. \\
\textbf{Output:} Carol knows if $w^Tx' > 0$.
\begin{enumerate}
\item Bob encrypts $w$ and transfers $E[w]$ to Carol.
\item Carol homomorphically computes the encrypted inner product.
  \[ \prod_{j=1}^d E[w]^{x_j'} 
  = E\left[\sum_{j=1}^d w_jx_j'\right] = E\left[w^Tx'\right]. \]
\item Carol generates a random number $r$ and sends
  $E\left[w^Tx'\right]-r$ to Bob.
\item Bob decrypts it to obtain his additive share $w^Tx'-r$. Let us
  denote it by $-s$, so that $r - s = w^Tx'$.
\item Bob and Carol execute a variant of the secure millionaire
  protocol~\cite{Yao82} to with inputs $r$ and $s$ and both learn
  whether $r > s$. \\
  If $r > s$, Carol concludes $w^Tx' > 0$ and if $r < s$, she
  concludes $w^Tx' < 0$.
\end{enumerate}

In this way, Carol and Bob are able to perform the classification
operation while maintaining the privacy constraints. If Bob has to
repeatedly execute the same protocol, he can
pre-compute $E[w]$ to be used in Step 1.

%% file: analysis.tex
\section{Analysis}

\subsection{Correctness}\label{sec:correctness}

The private training protocol does not alter any of the computations
of the original training algorithm and therefore results in the same
output. The additive randomization $r_i$ introduced in Step 3 is
removed in Step 6 leaving the results unchanged. Similarly, the
multiplicative randomization $q_i$ introduced in Step 7 is removed in
Step 9.

As discussed in Section~\ref{sec:homomorphic-encryption}, the only
source of error is the truncation of less significant digits in the
finite precision representation of real numbers. In practice, we
observe that the  error in computing the weight vector $w$ is
negligibly small and does not result in any loss of accuracy.
 
\input{security}

\subsection{Complexity}\label{sec:complexity}

We analyze the encryption/decryption and the data transmission 
costs for a single execution of the protocol as these consume a vast
majority of the time.

There are 6 steps of the protocol where encryption or decryption
operations are carried out. 
\begin{enumerate}
\item In Step 1, Bob encrypts the $d$-dimensional vector $w_{(t)}$.
\item In Step 3, Alice encrypts the $n$ random numbers $r_i$.
\item In Step 4, Bob decrypts the $n$ inner products obtained from Alice.
\item In Step 5, Bob encrypts the exponentiation of the $n$ inner products.
\item In Step 8, Bob decrypts, takes a reciprocal, and encrypts the
  $n$ multiplicatively scaled quantities.
\item In Step 12, Bob decrypts the $d$ dimensional updated weight
  vector obtained from Alice.
\end{enumerate}
\noindent \textbf{Total:} $3n + 2d$ encryptions and decryptions.~\\

Similarly, there are 6 steps of the protocol where Alice and Bob
transfer data to each other.
\begin{enumerate}
\item In Step 1, Bob transfers the $d$-dimensional vector
  $w_{(t)}$ to Alice.
\item In Step 3, Alice transfers $n$ randomized innner products to Bob.
\item In Step 5, Bob transfers the $n$ encrypted exponentials to Alice.
\item In Step 7, Alice transfers $n$ scaled quantities to Bob.
\item In Step 8, Bob transfers the $n$ encrypted reciprocals to Alice.
\item In Step 11, Alice transfers the $d$ dimensional encrypted
  updated weight vector to Bob.
\end{enumerate}
\textbf{Total:} Transmitting $4n + 2d$ elements.~\\

The speed of performing the encryption and decryption operations
depends directly on the size of the key of the
cryptosystem. Similarly, when we are transfering encrypted data, the
size of an individual element also depends on the size of the
encryption key. As the security of the encryption function is largely
determined by the size of the encryption key, this reflects a direct
trade-off between security and efficiency.

%% file: security.tex
\subsection{Security}\label{sec:security}

The principal requirement of a valid secure multiparty computation
(SMC) protocol is that any party must not learn anything about the
input data provided by the other parties apart from what can be
inferred from the result of the computation itself. As we mentioned
earlier, we assume that the parties are semi-malicious. From this
perspective, it can be seen that the private training protocol
(Section~\ref{sec:training-protocol}) is demonstrably secure. ~\\

\noindent{\em Alice/Carol:} In the private training protocol, Alice can only
observe encrypted inputs from Bob and hence she does not learn
anything about the weight vector used by Bob. In the private
classifier evaluation protocol, the party Carol with the test email
only receives the final outcome of the classifier in plaintext.
Thus, the only additional information available to her is the
output of the classifier itself, which being the output is permissible
under the privacy criteria of the problem.~\\

\noindent{\em Bob:} In the training stage, Bob receives unencrypted data
from Alice in Steps 3, 8 and 12.

\begin{itemize}
\item{} {\em Step 3}: Bob receives $y_i w^T_{(t)}x_i - r_i$. Let us
  denote this quantity by $v$ and $yw^Tx$ by $z$, giving us $v = z -
  r_i$. Since $r_i$ is drawn from a uniform distribution over the
  entire finite field $\mathbb{Z}_N$, for any $v$ and for every value
  of $z$ there exists a unique value of $r_i$ such that $v = z -
  r_i$. 
  Thus, $P_z(z|v) \propto P_z(z)P_r(z-v) = P_z(z)$.\footnote{The
    notation $P_x(X)$ denotes the probability with which the random
    variable $x$ has the value $X$.}
  The conditional entropy $H(z|v) = H(z)$, {\em i.e.}, Bob receives no
  information from the operation.

\item{}{\em Step 8}: A similar argument can be made for this step.
  Here Bob receives $v = q z$, where $z = 1 + e^{y_i w^T_{(t)} x_i}$.
  It can be shown that for any value $v$ that Bob receives,
  $P_z(z|v) \propto \frac{P_z(z) P_q(v/z)}{z}$. Since $q$ is drawn
  from a power law distribution, {\em i.e.} $P_q(q) \propto 1/q$,
  for all $v < |\mathcal{D}|$, $P_z(z|v) = P_z(z)$.
  Once again, the conditional entropy $H(z|v) =
  H(z)$, {\em i.e.}, Bob receives no information from the operation.

\item{} {\em Step 12}: The information Bob receives in this step is
  the updated weight vector, which is the result of the computation
  that Bob is permitted to receive by the basic premise of the SMC
  protocol.
\end{itemize}

\subsubsection*{Information Revealed by the Output}

We assume that all the parties agree with Bob receiving the
updated classifier at the end of the training protocol, this forms the
premise behind their participation in the protocol to start with.
If the parties use the modified training protocol which results in a
differentially private classifier, no information about the data can
be gained from the output classifier. In case the parties use the
original training protocol, the output classifier does reveal
information about the input data, which we quantify and present ways
to minimize in the following analysis.

At the end of Step 12 in each iteration, Bob receives the update
weight vector $w_{t+1} = w_t + \eta \nabla L(w_{(t)},x,y)$. As he also
has the previous weight vector $w_t$, he effectively observes the
gradient $\nabla L(w_{(t)},x,y) = \sum_i y_ix^T_i
\left(1+e^{y_iw^T_{(t)}x_i}\right)^{-1}$.

In the online setting, we normally use one training data instance at a
time to update the classifier. If Alice participates in the training
protocol using only one document $(x_1,y_1)$, the gradient observed by
Bob will be $y_1x_1\left(1+e^{y_1w^T_{(t)}x_i}\right)^{-1}$, which is
simply a scaling of the data vector $y_1x_1$. As Bob knows
$w_{(t)}$ he effectively knows $y_1x_1$. In particular, if $x_1$ is a
vector of non-negative counts as is the case for n-grams, the
knowledge of $y_1x_1$ is equivalent to knowing $x_1$. Although the protocol
itself is secure, the output reveals Alice's data completely.

Alice can prevent this by updating the classifier
using {\em blocks} of $K$ document vectors $(x,y)$ at a time.
The protocol ensures that for each block of $K$ vectors Bob only receives
the gradient computed over them
\begin{align*}
  \nabla L(w_{(t)},x,y) &= \sum_{i=1}^K y_ix^T_i \left(1+e^{y_iw^T_{(t)}x_i}\right)^{-1} \\
  &= \sum_{i=1}^K g(w_{(t)},x_i,y_i)x_i,
\end{align*}
where $g(w_{(t)},x_i,y_i)$ is a scalar function of the data instance
such that $g(w_{(t)},x_i,y_i)x_i$ has a one-to-one mapping to $x_i$.
Assuming that all data vectors $x_i$ are i.i.d., using
Jensen's inequality, we can show that the conditional entropy
\begin{align}\label{eqn:info-loss}
 H\left[x_i | \nabla L(w_{(t)},x,y)\right] \le \frac{K-1}{K} H[x_i] + \log(K).
\end{align}
In other words, while Bob gains some information about the data
belonging to Alice, the amount of this information is inversely
proportional to the block size. In the online learning setting,
choosing a large block size decreases the accuracy of the classifier.
Therefore, the choice of the block size effectively becomes a
parameter that Alice can control to trade off giving away some
information about her data with the accuracy of the classifier. In
Section 6.2, we empirically analyze the performance of the classifier
for varying batch sizes. We observe that in practice, the accuracy of
the classifier is not reduced even after choosing substantially large
batches of 1000 documents, which would hardly cause any loss of
information as given by Equation~\ref{eqn:info-loss}.

%% file: experiments.tex
\section{Experiments}\label{sec:experiments}

We provide an experimental evaluation of our approach for the task of
email spam filtering. The privacy preserving training protocol
requires a substantially larger running time as compared to the
non-private algorithm. In this section, we analyze the training
protocol for running time and accuracy. As the execution of the
protocol on the original dataset requires an infeasible amount of
time, we see how data independent dimensionality reduction can be
used to effectively reduce the running time while still
achieving comparable accuracy.

As it is conventional in spam filtering research, we report AUC
scores.\footnote{Area under the ROC curve.} It is considered to be a
more appropriate metric for this task as compared to other metrics
such as classification accuracy or F-measure because it averages the
performance of the classifier in different precision-recall points
which correspond to different thresholds on the prediction confidence
of the classifier. The AUC score of a random classifier is 0.5 and
that for the perfect classifier is 1. We compared AUC performance of
the classifier given by the privacy preserving training protocol with
the non-private training algorithm and in all cases the numbers were
identical up to the five significant digits. Therefore, the error due
to the finite precision representation mentioned in
Section~\ref{sec:correctness} is negligible for practical
purposes.

\begin{table}
\centering
\caption{Email spam dataset summary.}
\label{tab:dataset}
\begin{tabular}{|c|c|c|c|} \hline
  Section  & Spam        & Non-spam   & Total \\ \hline
  Training & 2466 (82\%) & 534 (18\%) & 3000 \\ \hline
  Testing  & 2383 (79\%) & 617 (21\%) & 3000 \\ \hline
\end{tabular}
\end{table}

\subsection{Email Spam Dataset}

We used the public spam email corpus from the CEAS 2008 spam filtering
challenge.\footnote{The dataset is available at
  \url{http://plg.uwaterloo.ca/~gvcormac/ceascorpus/} The part of the
  dataset we have used corresponds to \texttt{pretrain-nofeedback}
  task.} For generality, we refer to emails as documents. Performance
of various algorithms on this dataset is reported in
\cite{emailspam}. The dataset consists of 3,067 training and 206,207
testing documents manually labeled as spam or ham (\emph{i.e.}, not
spam). To simplify the benchmark calculations, we used the first 3000
documents from each set (Table~\ref{tab:dataset}). Accuracy of the
baseline majority classifier which labels all documents as spam is
0.79433.

\subsection{Spam Filter Implementation}\label{sec:filterimpl}

Our classification approach is based on online logistic regression
\cite{Goodman06}, as described in Section \ref{sec:lr}. The features are
overlapping character four-grams which are extracted from the documents
by a sliding window of four characters. The feature are binary
indicating the presence or absence of the given four-gram. The documents
are in \texttt{ASCII} or \texttt{UTF-8} encoding which
represents each character in 8 bits, therefore the space of possible
four-gram features is $2^{32}$. Following the previous work, we used
modulo $10^{6}$ to reduce the four-gram feature space to one million
features and only the first 35 \texttt{KB} of the documents is used to
compute the features. For all experiments, we use a step size of
$\eta=0.001$ and no regularization or noise required for differential
privacy is used.

\begin{table}[ht]
  \centering
  \caption{Running time comparison of online training of logistic
    regression (LR) and the privacy preserving logistic regression
    (PPLR) for one document.}
  \label{tab:speed}
  \begin{tabular}{|c|r|r|} \hline
    Feature Count & LR & PPLR\\ \hline
    Original: $10^{6}$ & 0.5 s  & 1.14 hours \\ \hline
    Reduced: $10^{4}$  & 5 ms   & 41 s \\ \hline
  \end{tabular}
\end{table}

\begin{table}[ht]
  \centering
  \caption{Running time of privacy preserving logistic regression for one document of $10^4$ features with different encryption key sizes.}
  \label{tab:encspeed}
  \begin{tabular}{|c|r|} \hline
    Encryption Key Size & Time\\ \hline
    256 bit  & 41 s \\ \hline
    1024 bit & 2013 s\\ \hline
  \end{tabular}
\end{table}

\begin{table}[ht]
  \centering
  \caption{Time requirement for steps of the protocol for random matrices of the dimensions shown (documents $\times$ features).}
  \label{tab:steptime}
  \begin{tabular}{|c|r|r|} \hline
    Steps&Time (s) - 200$\times$20&Time (s) - 200$\times$100\\ \hline
    1&0.06&0.31\\ \hline
    2, 3&2.59&10.14\\ \hline
    4, 5&0.82&0.73\\ \hline
    6, 7&0.46&0.41\\ \hline
    8&0.84&0.73\\ \hline
    9, 10&1.81&8.33\\ \hline
    11&0.05&0.18\\ \hline\hline
    Total&6.61&20.81\\ \hline
  \end{tabular}
\end{table}

\subsection{Protocol Implementation}

We created a prototype implementation of the protocol in C++ and used
the variable precision arithmetic libraries provided by
OpenSSL~\cite{openssl} to implement the Paillier cryptosystem.  We
used the GSL libraries~\cite{gsl} for matrix operations.  We performed
the experiments on a 3.2 GHz Intel Pentium 4 machine with 2 GB RAM and
running 64-bit Ubuntu.

The original dataset has $10^6$ features as described in
Section~\ref{sec:filterimpl}. Similar to the complexity analysis of
the training protocol (Section~\ref{sec:correctness}), we
observed that time required for the training protocol is linear in
number of documents and number of features. 

Table~\ref{tab:speed} compares the time required to train a logistic
regression classifier with and without the privacy preserving protocol
using 256-bit encryption for one document. It can be seen that the
protocol is slower than non-private version by a factor of $10^4$
mainly due to the encryption in each step of the protocol. Also, we
observe that the running time is drastically reduced with the
dimensionality reduction. While the execution time for the
training protocol over the original feature space would be infeasible
for most applications, the execution time for the reduced feature
space is seen to be usable in spam filtering
applications. This motivated us to consider various dimensionality
reduction schemes which we discuss in Section \ref{sec:dimensionality}.

To further analyze the behavior of various steps of the protocol, in
Table~\ref{tab:steptime} we report the running time of individual
steps of the protocol outlined in Section~\ref{sec:training-protocol}
on two test datasets of random vectors. It can be observed that
encryption is the main bottle neck among the other operations in the
protocol.  We report the Paillier cryptosystem with 256-bit keys in
the following experiments.  As shown in Table~\ref{tab:encspeed},
using the more secure 1024-bit encryption keys, resulted in a slowdown
by a factor of about 50 as compared to using 256-bit encryption keys.
This is a constant factor which can be applied to all our timing
results if the stronger level of security provided by 1024-bit keys is
desired.

Using a pre-computed value of the encrypted weight vector $E[w]$, the
private evaluation protocol took 210.956 seconds for one document using
$10^6$ features and 2.059 seconds for one document using
$10^4$ features which again highlights the necessity for dimensionality
reduction to make the private computation feasible.

\subsection{Dimensionality Reduction} \label{sec:dimensionality}

Since the time requirement of the privacy preserving protocol varies
linearly with the data dimensionality, we can improve it
by dimensionality reduction principally because data with fewer number
of features will require fewer encryptions and decryptions.
On the other hand, reducing the dimensionality of the features, particularly for
 sparse features such as $n$-gram counts, can have an effect on the
 classification performance. We study this behavior by 
experimenting with six different dimensionality
reduction techniques, and compared the running time and AUC of the
classifier learned by the training protocol.

We consider PCA which is a data-dependent dimensionality reduction
technique and five other ones which are data independent. The latter
techniques are much more in our setting as they can be used
by multiple parties on their individual documents without violating
privacy.

\begin{table}[ht]
  \centering
  \caption{Performance of PCA for dimensionality reduction.}
  \label{tab:pca}
  \begin{tabular}{|c|r|r|} \hline
    Dimension&Time (s) & AUC~~ \\ \hline
    5&18&0.96159\\ \hline
    10&37&0.99798\\ \hline
    50&242&0.99944\\ \hline
    100&599&0.99967\\ \hline
    300&5949&0.99981\\ \hline
  \end{tabular}
\end{table}

\begin{table}
  \centering
  \caption{Time and space requirement for dimensionality reduction
    methods for reduction from $10^6$ to $10^4$ features.}
  \label{tab:drtimespace}
  \begin{tabular}{|c|c|c|} \hline
    Method  & Time (s)& Space (GB) \\ \hline
    PCA     & 7 $\times 10^6$&41\\ \hline
    LSH     & 50 $\times 10^3$&40\\ \hline
    Hash Space & 41 & -- \\ \hline
    Document Frequency&1& -- \\ \hline
    Sample Uniform&2& -- \\ \hline
    Sample Multinomial&490& -- \\ \hline
  \end{tabular}
\end{table}

\begin{figure}[ht]
  \centering
  \includegraphics[width=\columnwidth]{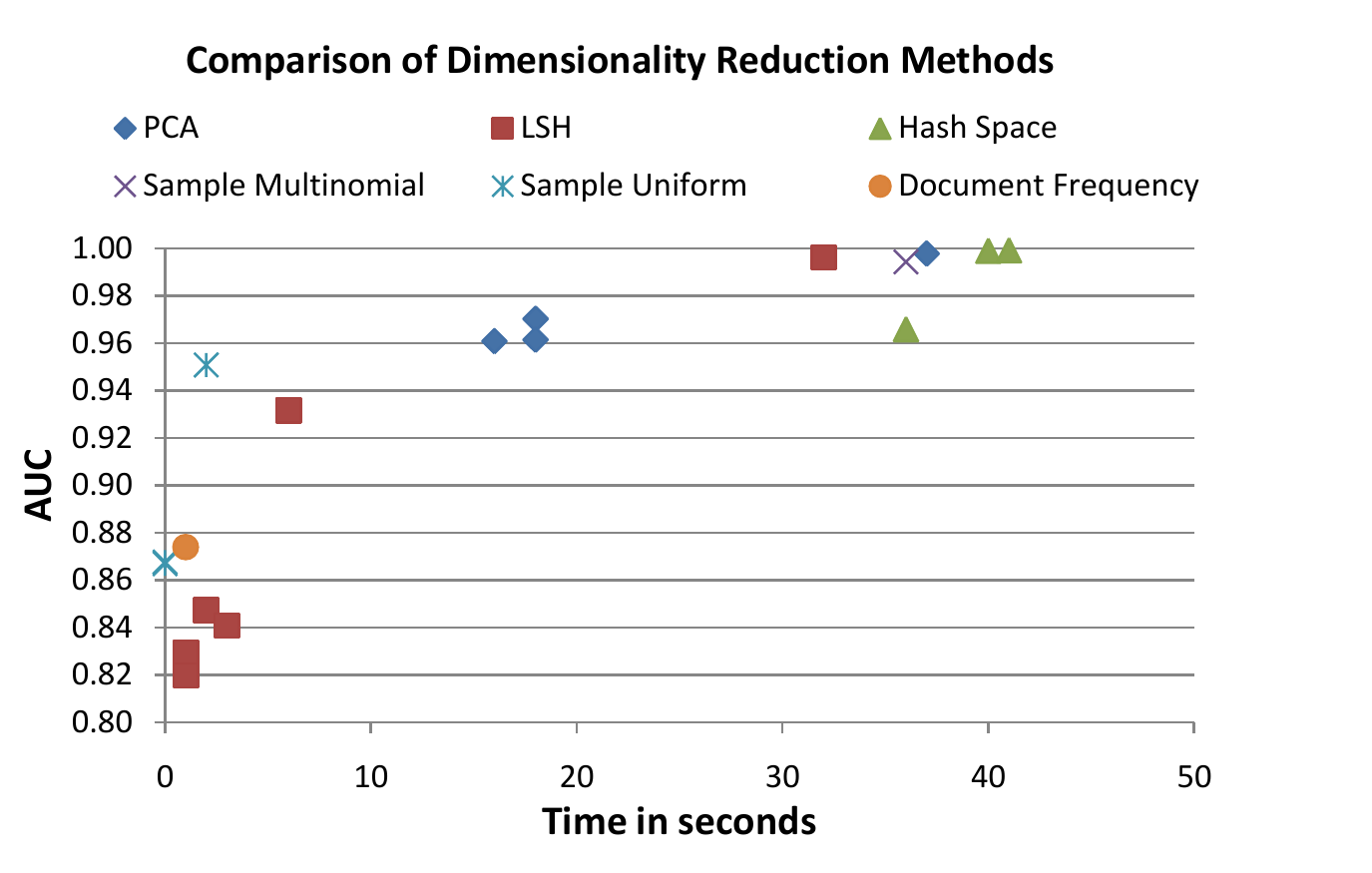}
  \caption{Time comparison for the dimensionality reduction approaches
    reduced from $10^6$ to $10^4$ dimensions.}
  \label{fig:timeauc}
\end{figure}

\begin{figure}[ht]
  \centering
  \includegraphics[width=.9\columnwidth]{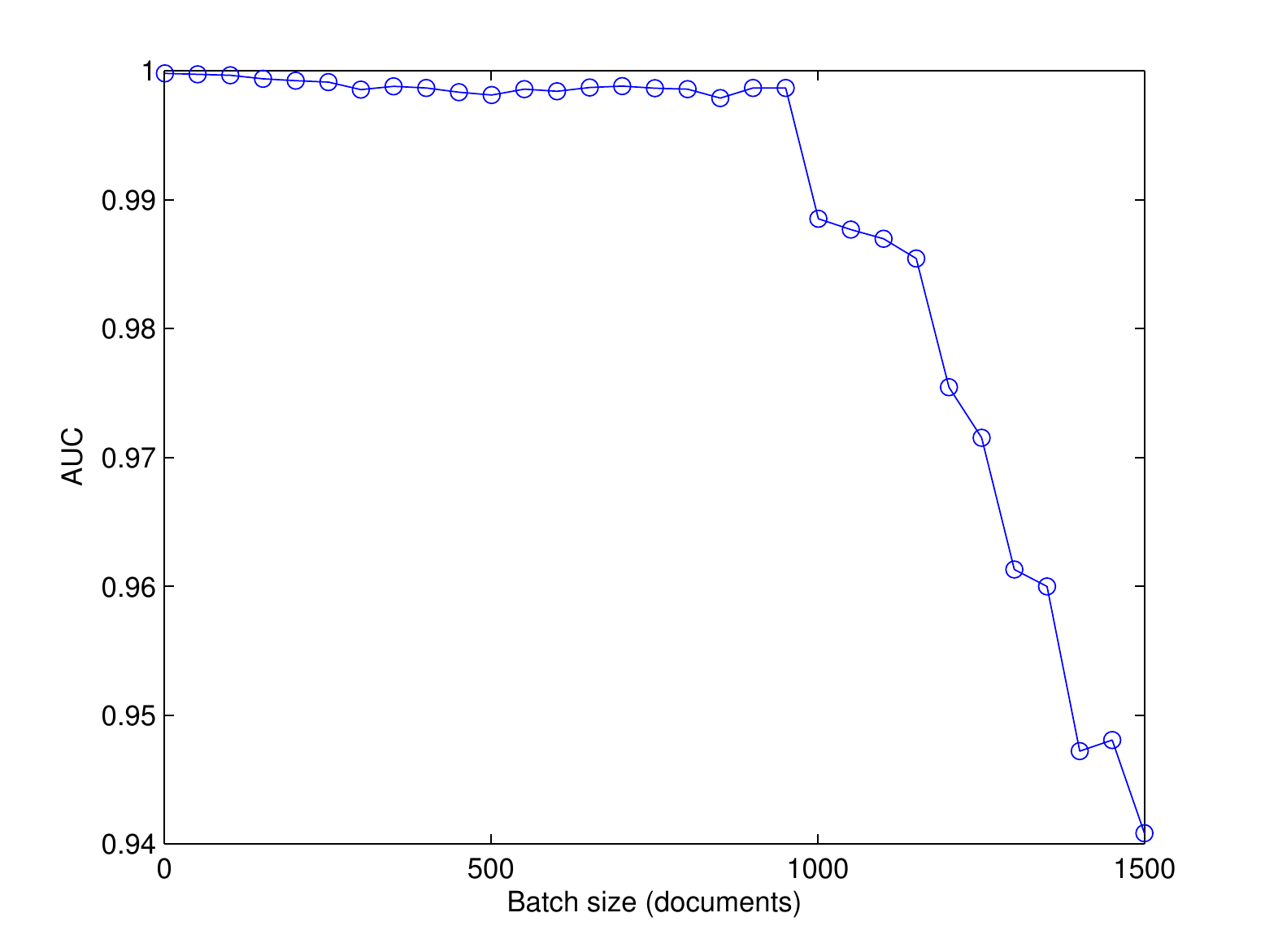}
  \caption{Performance of one iteration of logistic regression
    training on 300 dimensional PCA feature vectors with different
    batch sizes.}
  \label{fig:batching}
\end{figure}

\begin{enumerate}
\item {\bf Principal Component Analysis (PCA):} PCA is perhaps the
  most commonly used dimensionality reduction technique which computes
  the lower dimensional projection of the data based on the most dominant
  eigenvectors of covariance matrix of the original data. Since we
  only compute a small number of eigenvectors, PCA is found to be
  efficient for our sparse binary dataset. Table~\ref{tab:pca}
  summarizes the running time and the AUC of the classifier trained on
  the reduced dimension
  data.
  While the performance of PCA is excellent, it has the following
  disadvantages, motivating us to look at other techniques.

  \begin{enumerate}
  \item When training in a multiparty setting, all the parties are
    required to use a common feature representation. Among the methods
    we considered, only PCA computes a projection matrix which is data
    dependent. This projection matrix cannot be computed over the
    private training data because it reveals information about the
    data.

  \item For many classification tasks, reduction to an extremely small
    subspace hurts the performance much more significantly than in our
    case. Furthermore, computing PCA with high dimensional data is not
    efficient and we are interested in efficient and scalable
    dimensionality reduction techniques.
  \end{enumerate}

\item {\bf Locality Sensitive Hashing (LSH):} In LSH~\cite{lsh}, we
  choose $k$ random hyperplanes in the original $d$ dimensional space
  which represent each dimension in the target space. 
  The reduced dimensions are binary and indicate the side of the
  hyperplane on which the original point lies.

\item {\bf Hash Space Reduction:} As mentioned in Section \ref{sec:filterimpl},
  we reduce the original feature space to modulo $10^{6}$. We
  experimented with different sizes of this hash space.

\item {\bf Document Frequency Based Pruning:} We select
  features which occur in at least $k$ documents.
  This is a common approach in removing rarely-occurring features,
  although some of those feature could be discriminative especially in
  a spam filtering task.

\item {\bf Uniform Sampling:} In this approach, we draw from the
  uniform distribution until desired number of unique features are
  selected.

\item {\bf Multinomial Sampling:} This approach is similar to the
  uniform sampling approach except that we first fit a multinomial
  distribution based on the document frequency of the features and
  then draw from this distribution. This causes the sampling to be
  biased toward features with higher variance which are often the more
  informative features.

\end{enumerate}

We ran each of these algorithms on 6000 documents of $10^6$
dimensions. Table~\ref{tab:drtimespace} summarizes the time and space
requirement of each algorithm for reducing dimensions to $10^4$. We
trained the logistic regression classifier on 3000 training documents
with various reduced dimensions and measured the running time and AUC
of the learned classifier on the 3000 test documents. The results are
shown in Figure~\ref{fig:timeauc}. We observe that the data
independent dimensionality reduction techniques such as LSH,
multinomial sampling, and hash space reduction achieve close to
perfect AUC.

\subsubsection*{Classifier Performance for Varying Batch Size}

As we discussed in Section~\ref{sec:security}, another important
requirement of our protocol is to train in batches of documents rather than
training on one document at a time. We have shown
that the extra information gained by Bob about any party's data decreases
with the increasing batch size. On the other hand, increasing the batch
size causes the optimization procedure of the training algorithm
to have fewer chances of correcting itself in a single
pass over the entire training dataset. In Figure~\ref{fig:batching},
we see that the trade-off in AUC is negligible even with batch sizes
of around 1000 documents.

\subsection{Parallel Processing}

An alternative approach to address the performance issue is
parallelization. We experimented with a multi-threaded implementation
of the algorithm. On average, we observed 6.3\% speed improvement on a
single core machine. We expect the improvement to be more
significant on a multi-core architecture. A similar scheme can
be used to parallelize the protocol across a cluster of machines, such as
in a MapReduce framework. In both of these cases, the accuracy of the
online algorithms will decrease slightly as the number of threads or
machines increase because the gradient $\nabla L(w_{(t)},x,y)$
computed in each of the parallel processes is based on an older value
of the weight vector $w_{(t)}$.

A more promising approach which does not impact the accuracy is
encrypting vectors in parallel. In the present implementation of the
protocol, we encrypt vectors serially and the procedure used
for the individual elements is identical. We can potentially
reduce the encryption time of a feature vector substantially by using
a parallel processing infrastructure such as GPUs. We leave the
experiments with such an implementation for future work.

%% file: conclusion.tex
\section{Conclusion}

We developed protocols for training and evaluating a logistic
regression based spam filtering classifier over
emails belonging to multiple parties while preserving the privacy
constraints. 
We presented an information theoretic
analysis of the security of the protocol and also found that both the
encryption/decryption and data transmission costs of the protocol are
linear in the the number of training instances and the
dimensionality of the data. We also experimented with a prototype
implementation of the protocol on a large scale email dataset and
demonstrate that our protocol is able to achieve close to state of the
art performance in a feasible amount of execution time.

The future directions of this work include applying our methods to
other spam filtering classification algorithms. We also plan to extend
our protocols to make extensive use of parallel architectures such as
GPUs to further increase the speed and scalability.

%% file: pplr.bbl
\begin{thebibliography}{10}

\bibitem{shingles}
A.~Z. Broder.
\newblock Some applications of {Rabin's} fingerprinting method.
\newblock {\em Sequences II: Methods in Communications, Security, and Computer
  Science}, pages 143--152, 1993.

\bibitem{lsh}
M.~Charikar.
\newblock Similarity estimation techniques from rounding algorithms.
\newblock In {\em 34th Annual {ACM} Symposium on Theory of Computing}, 2002.

\bibitem{Cormack07}
G.~V. Cormack.
\newblock {TREC} 2007 spam track overview.
\newblock In {\em Text REtrieval Conference {TREC}}, 2007.

\bibitem{gsl}
M.~Galassi, J.~Davies, J.~Theiler, B.~Gough, G.~Jungman, P.~Alken, M.~Booth,
  and F.~Rossi.
\newblock {\em GNU Scientific Library Reference Manual (v1.12)}.
\newblock Network Theory Ltd., third edition, 2009.

\bibitem{Goodman06}
J.~Goodman and W.~Yih.
\newblock Online discriminative spam filter training.
\newblock In {\em Conference on Email and Anti-Spam {CEAS}}, 2006.

\bibitem{Li2009}
K.~Li, Z.~Zhong, and L.~Ramaswamy.
\newblock Privacy-aware collaborative spam filtering.
\newblock {\em IEEE Transactions on Parallel and Distributed Systems},
  20(5):725--739, 2009.

\bibitem{LinCZ05}
X.~Lin, C.~Clifton, and M.~Y. Zhu.
\newblock Privacy-preserving clustering with distributed {EM} mixture modeling.
\newblock {\em Knowledge and Information Systems}, 8(1):68--81, 2005.

\bibitem{openssl}
\url{http://www.openssl.org/docs/crypto/bn.html}.

\bibitem{Paillier99}
P.~Paillier.
\newblock Public-key cryptosystems based on composite degree residuosity
  classes.
\newblock In {\em EUROCRYPT}, 1999.

\bibitem{emailspam}
D.~Sculley and G.~V. Cormack.
\newblock Going mini: Extreme lightweight spam filters.
\newblock In {\em Conference on Email and Anti-Spam {CEAS}}, 2008.

\bibitem{spam-report}
Symantec intelligence report: August 2011.
\newblock
  \url{http://www.symantec.com/connect/blogs/symantec-intelligence-report-augu%
st-2011}.

\bibitem{VaidyaCKP08}
J.~Vaidya, C.~Clifton, M.~Kantarcioglu, and S.~Patterson.
\newblock Privacy-preserving decision trees over vertically partitioned data.
\newblock {\em TKDD}, 2(3), 2008.

\bibitem{VaidyaKC08}
J.~Vaidya, M.~Kantarcioglu, and C.~Clifton.
\newblock Privacy-preserving naive {Bayes} classification.
\newblock {\em VLDB J}, 17(4):879--898, 2008.

\bibitem{jvppsvm}
J.~Vaidya, H.~Yu, and X.~Jiang.
\newblock Privacy-preserving {SVM} classification.
\newblock {\em Knowledge and Information Systems}, 14(2):161--178, 2008.

\bibitem{Yao82}
A.~Yao.
\newblock Protocols for secure computations.
\newblock In {\em IEEE Symposium on Foundations of Computer Science}, 1982.

\end{thebibliography}
